\documentclass[11pt,a4paper]{article}
\usepackage[hyperref]{acl2020}
\usepackage{times}
\usepackage{latexsym}

\usepackage{microtype}

\usepackage{times}
\usepackage{latexsym}
\usepackage{graphicx}
\usepackage{url}
\usepackage{enumitem}
\usepackage{booktabs}
\usepackage{xspace}
\usepackage[normalem]{ulem}

\newcommand{\wctan}[1]{{\it\small\textcolor{blue}{[[[ {#1}\ --wangchiew ]]]}}}

\newcommand{\yoshi}[1]{{\it\small\textcolor{teal}{[[[ {#1}\ --yoshi]]]}}}

\newcommand{\xsensekb}{{\sc xSense KB}\xspace}
\newcommand{\systemnamecn}{{\sc xSense(CN)}\xspace}
\newcommand{\hsensebase}{{\sc BERT+SQuAD}\xspace}
\newcommand{\squad}{{\sc SQuAD {\small v}1.1}\xspace}
\newcommand{\systemname}{{\sc xSense}\xspace}

\newcommand{\spara}[1]{\smallskip\noindent{\bf{#1}}}
\newcommand{\mpara}[1]{\medskip\noindent{\bf{#1}}}

\newcommand{\phrase}[1]{``{\em #1}''}

\title{Enhancing Review Comprehension with Domain-Specific Commonsense}

\author{Aaron Traylor \\
  Brown University \\
  \texttt{aaron\_traylor@brown.edu} \\\And
  Chen Chen \\\And
  Behzad Golshan \\
  Megagon Labs \\
  \texttt{\{chen, behzad, xiaolan\}@megagon.ai} \\\And
  Xiaolan Wang \\\AND
  Yuliang Li \\\And
  Yoshihiko Suhara \\\And
  Jinfeng Li \\
  Megagon Labs \\
  \texttt{\{yuliang, yoshi, jinfeng, cagatay, wangchiew\}@megagon.ai} \\\And
  Cagatay Demiralp \\\And
  Wang-Chiew Tan}

\date{}

\begin{document}
\maketitle

\begin{abstract}
Review comprehension has played an increasingly important role in improving the quality of online services and products and commonsense knowledge can further enhance review comprehension.
However, existing general-purpose commonsense knowledge bases lack sufficient coverage and precision to meaningfully improve the comprehension of domain-specific reviews. In this paper,  we introduce \systemname, an effective system for review comprehension using domain-specific commonsense knowledge bases ({\xsensekb}s). We show that {\xsensekb}s can be constructed inexpensively and present a knowledge distillation method that enables us to use {\xsensekb}s along with BERT to boost the performance of various review comprehension tasks. We evaluate \systemname over three review comprehension tasks: aspect extraction, aspect sentiment classification, and question answering. We find that \systemname outperforms the state-of-the-art models for the first two tasks and improves the baseline BERT QA model significantly, demonstrating the usefulness of incorporating commonsense into review comprehension pipelines. To facilitate future research and applications, we publicly release three domain-specific knowledge bases and a domain-specific question answering benchmark along with this paper.
\end{abstract}

\section{Introduction}




Today, many consumer services have significant online presence which makes it easy for consumers to leave feedback and reviews on the services rendered. 
Various popular NLP tasks are relevant to review comprehension,
including aspect extraction (AE),
aspect sentiment classification (ASC), and question answering (QA). 
Despite the progress made on these fronts, holistic understandings of reviews' meanings often requires commonsense reasoning by the reader. 
%
In recent years, several pre-training techniques~\cite{Devlin:2019:BERT,Liu:2019:RoBERTa,Yang:2019:XLNet} have shown the state-of-the-art (SOTA) performance on commonsense reasoning
tasks~\cite{Zellers:2018:SWAG,Talmor:2019:CommonsenseQA,Huang:2019:CosmosQA,Zhang:2018:ReCoRD}.
However, these solutions are still inadequate in general for review
comprehension as different domains tend to adopt languages and commonsense that
are domain-specific. 
For example, the hotel review ``{\em The
place is 800m away from the beach!}'' conveys positive information about its walking distance, convenience, and location and can be used to answer questions such as
whether the hotel is close to the beach, whether it is within walking distance,
or whether it is in a desirable location. It will be difficult to answer these questions
without domain-specific commonsense.

Table~\ref{table:kbexample} shows more examples of the type of commonsense that would be useful for accurately interpreting reviews in different domains.
As our experiments show,
these types of commonsense cannot be derived from popular
commonsense knowledge bases such as ConceptNet~\cite{liu2004Conceptnet}, which
also yields sub-optimal results for review comprehension tasks when compared to using our collected
domain-specific commonsense.

\begin{table}[!ht]
  \centering
  \small
  \begin{tabular}{@{}lcc@{}}
  \toprule  
  \textbf{Domain} & \textbf{Premise} & \textbf{Conclusion}  \\
  \midrule
  \midrule
  Hospitality & thin walls & noisy room\\
  Hospitality & near beach & good location\\
  Restaurants & good music & great vibe\\
  Restaurants & rude hostess & poor service\\
  Laptops & rich bass & crisp sound\\
  Laptops & excellent contrast & good image\\
  \bottomrule
  \end{tabular}
  \caption{Examples of domain-specific commonsense.}
  \label{table:kbexample}
\end{table}



More specifically, our contributions are:
\vspace{-2mm}
\begin{itemize}[leftmargin=*,noitemsep]
    \item We developed \systemname, a system that leverages domain-specific commonsense 
    knowledge bases (KBs) to
    enhance BERT~\cite{Devlin:2019:BERT} for various review-reading comprehension tasks such as aspect extraction,
    aspect sentiment classification, and question answering.     
    \item We present a method to collect and organize domain-specific commonsense KBs
    with relatively low cost. Less than \$$700$ was spent for each domain to collect
    KBs with roughly $6,000$ commonsense facts.
    \item We show that
    \systemname\ consistently achieves competitive or SOTA performance for multiple
    review comprehension tasks with relatively small commonsense KBs across 3 different domains.
    Specifically, we gain $1.5$ absolute F1 improvement for the QA task and outperform
    the SOTA models by up to $2.42$ F1 and $3.18$ Macro-F1 for the AE and ASC tasks respectively. 
    \item To facilitate future research, 
    we release three domain-specific KBs in the \emph{hospitality},
    \emph{restaurant} and \emph{laptop} domains. We
    also release an adversarial domain-specific question answering benchmark for the hospitality domain. 
\end{itemize}

The rest of the paper is organized as follows. Section~\ref{sec:overview}
provides an overview of \systemname. Its architecture is described in Section~\ref{sec:xsense} and we discuss our \xsensekb construction method in Section~\ref{sec:kb}. We demonstrate the 
advantages of using {\xsensekb}s in our pipeline in our experiments in Section~\ref{sec:experiment}. Finally, we discuss related work in
Section~\ref{sec:related} and conclude the paper in
Section~\ref{sec:conclusion}.

\section{System Overview}
\label{sec:overview}
\begin{figure}
    \centering
    \includegraphics[scale=0.4]{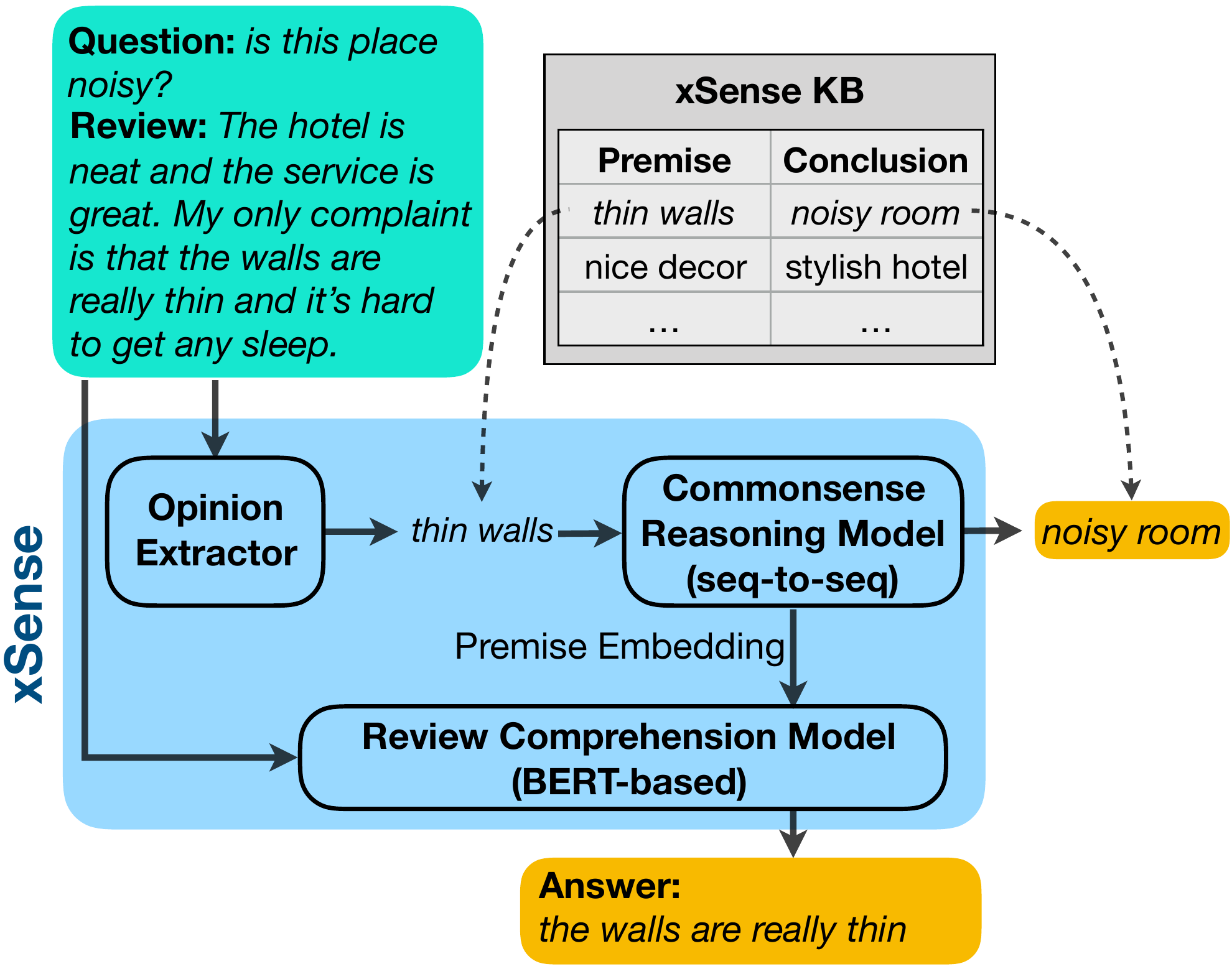}
    \caption{Overview of \systemname's architecture.}
    \label{fig:architecture}
\end{figure}

\systemname (Figure~\ref{fig:architecture}) takes  a question and a review as input and
returns a single span as the answer to the question. 
The architecture of \systemname has three main components: (1) an opinion extractor, (2) a commonsense reasoning model, and
(3) a review-comprehension model.
The same architecture can be used for other
reading comprehension tasks such as aspect extraction and aspect sentiment classification. The input and output for other 
tasks are different and minor adjustments are required as we explain in
Section~\ref{sec:xsense}.

The opinion extractor is responsible for extracting spans of the input reviews
that convey the reviewers' opinions, such as \phrase{tasty sushi}
and \phrase{short battery life}. 
The opinion extractor extracts such spans of opinions from
the input review and forwards them to the commonsense reasoning model to figure out what each 
extracted opinion entails. In our implementation, we use the opinion extractor from
OpineDB~\cite{opinedb}, which is the state-of-the-art tool for opinion mining
from reviews.

The pre-trained commonsense reasoning model identifies what conclusions
can be derived from the extracted opinions. For instance, \phrase{tasty sushi} 
might imply a \phrase{good Japanese restaurant}, and \phrase{short battery life}
implies \phrase{poor quality}. We refer to the input extraction as a
\emph{premise} and the output of the commonsense reasoning model as a
\emph{conclusion}. The commonsense reasoning model has been trained to identify correct
conclusions in a pre-training phase using available {\xsensekb}s. In addition to a conclusion, the commonsense
reasoning model also outputs an embedding for each premise. These embeddings encode
the knowledge that the model has for input premises
and can be used to
enhance the performance of reading comprehension tasks. 

Finally, 
the review comprehension model uses BERT to compute a 
representation for the input text. This representation is then augmented with
the premise embeddings from the commonsense reasoning model to further enhance
the output of the review comprehension model which, in this case, corresponds to identifying the answer span.

In short, the \systemname\ pipeline is effective for (1) identifying
the parts of texts that are good candidates for commonsense reasoning, (2) predicting what each
extracted span from the review entails and encodes this knowledge in an
embedding vector, and (3) using these embedding vectors along with BERT to
produce better results. 


\section{XSense}
\label{sec:xsense}
In this section, we detail each component in \systemname with the assumption that an \xsensekb\ is available for the desired domain.

\subsection{Opinion Extractor}
The opinion extractor 
takes a review as input and
outputs 
opinion tuples in the schema of (modifier, aspect). For
example, given a review ``{\em The bathroom is very clean but the food is
average.}'', the extractor would extract $\{$(very clean, bathroom),
(average, food)$\}$.
The extraction pipeline in \cite{opinedb} leverages two models: 
a sequence tagging model to identify the aspect and modifier spans and
a sequence pair classifier to combine aspects with their corresponding modifiers.

\subsection{Commonsense Reasoning Model}
The goal of the commonsense
reasoning model is to predict what \emph{conclusions} can be derived from the input
\emph{premise} given as a (modifier, aspect) pair. 
This is done by creating an embedding for each input premise to encode
the possible conclusions the input entails. Note that since conclusions
are derived from these embeddings, premises with similar conclusions
tend to have similar embeddings.

To obtain these premise embeddings, our reasoning model follows a standard
sequence-to-sequence model, with a 50-dimensional embedding layer and a 768-dimensional hidden layer of a gated recurrent unit (GRU)~\cite{cho2014properties} for both the encoder and the decoder. The embedding layer is initialized with GloVe word embeddings~\cite{pennington2014glove}. 
Given an
\xsensekb, which follows the schema shown in Table~\ref{table:kbexample},
we train the model with each premise-conclusion pair as a pair of
input-output sequences. 

Note that there are many techniques for knowledge-base embedding~\cite{yang2015embedding,nickel2012factorizing,trouillon2016complex} which in theory
could have been used in \systemname\ to embed the commonsense knowledge. However, these 
techniques only compute embeddings for entities that are present in the knowledge-base
and cannot generalize beyond those entities. In our case, entities in the knowledge
base are opinions expressed in natural language form, and thus by using a sequence-to-sequence
model, we can generalize beyond what appears in our knowledge-base. For instance, even if the
phrase \phrase{fresh nigiri} does not appear in the \xsensekb\ for the restaurant domain, our
approach infers that this premise implies \phrase{good Japanese restaurant}
because the phrase \phrase{fresh sashimi} has the same conclusion, and there is a high degree
of similarity between the two premises according to word embeddings.

\subsection{Review Comprehension Model}
The review comprehension model extends BERT to incorporate the embeddings obtained from
the commonsense reasoning model. In what follows, we 
overview BERT's architecture for each review comprehension task and explain
how the embeddings are utilized by \systemname.

\begin{figure}
    \centering
    \includegraphics[scale=.33]{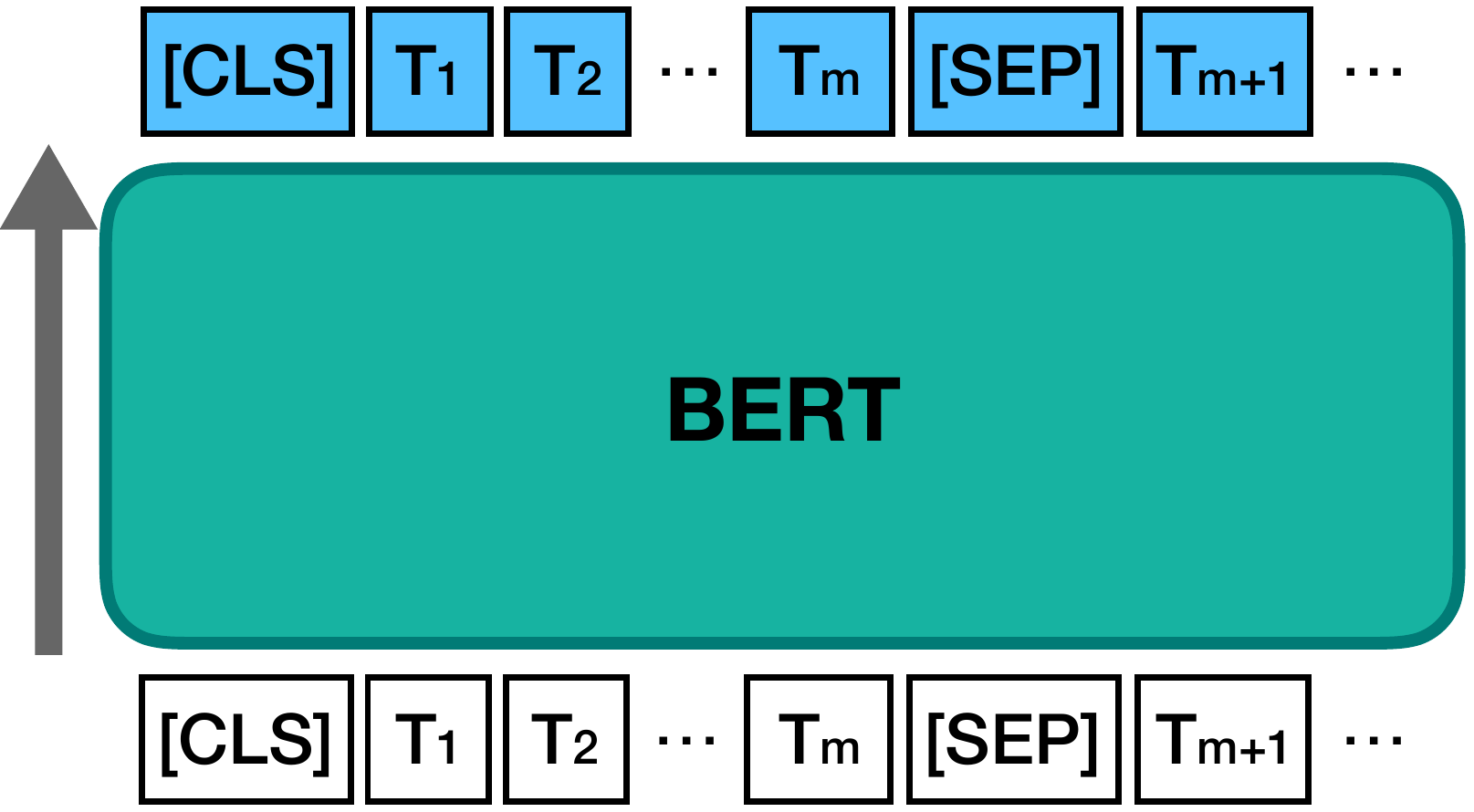}
    \caption{Overview of BERT representations}
    \label{fig:bert}
\end{figure}

\mpara{Overview.}
Figure~\ref{fig:bert} shows how BERT processes the input text to produce a representation
for each token (shown in blue) as well as the entire provided text. Note that for each token
$T_i$ in the input text, BERT outputs a representation (of size 768 or 1024). 
Besides the tokens
present in the text, BERT uses two special tokens \texttt{CLS} which is used to encode
a representation for the entire input text and \texttt{SEP} which is used to signal BERT
about specific aspect of the task at hand. For instance, for question answering tasks
the \texttt{SEP} token is used to separate the tokens of the input question from the 
tokens of the input text.

To use BERT for different NLP tasks, a final layer\footnote{Often a single dense layer}
is added on top of the learned representations. For instance, for sentiment analysis, a single dense 
layer is added on top of the \texttt{CLS} token which predicts the sentiment of the input text.

The \xsensekb\ follows the same approach, but augments the BERT representations with
embeddings obtained from the commonsense reasoning model. More specifically, the
representation of each token $T_i$ from a sentence $s$ is appended with the embedding
of an opinion extracted from sentence $s$. If the opinion extractor has no
opinions mined from sentence $s$, a vector of all zeros is appended instead.
If there are multiple opinions extracted from a sentence, we simply pick the
first extraction and append its embedding.

\mpara{Aspect Extraction.}
This task identifies the tokens in a given 
review that are aspects of the item or the service being reviewed. For instance,
\phrase{food} in the review ``{\em The food was tasty, but ...}'' is an aspect
to be extracted. The input to this task is the \texttt{CLS} token followed
by the tokens of the review.
To predict which tokens should be extracted, 
a single dense layer is added on top of the BERT representations which are
augmented by adding the commonsense embeddings to these representations.
The dense layer outputs the probability of whether  or not each token is part of an
aspect span. 

\mpara{Aspect Sentiment Classification.} 
The input to the aspect sentiment classifier is a review along with a
span marked in the review as the targeted aspect. The goal is to predict
whether the reviewer's opinion on the aspect is positive, negative, or neutral. 
The input is provided to BERT in the same manner as the aspect-extraction
task with one minor adjustment: the targeted aspect is appended to the original review
after a \texttt{SEP} token. To predict the
expressed sentiment, a dense layer is often added on top of the \texttt{CLS}
token which is fine-tuned during the training process. However, \systemname\
augments the \texttt{CLS} representation by adding the commonsense embedding
of the input text. A dense layer is added on top of this augmented representation
to make the final prediction.

\mpara{Question Answering.}
Given a question and a review (which is assumed to contain the answer to
the question), the goal is to find the
span in the review that can be served as the answer to the question.
This input is fed to BERT by separating the question and the review using
a \texttt{SEP} token.

To identify which span has the highest likelihood of being the correct answer, two single dense-layer classifiers are added on top of the BERT representations
of each token appended with their associated commonsense embeddings. The 
two classifiers compute the likelihood of each token being at the start and at
the end of the answer span respectively. Based on these probabilities the
span with the highest likelihood of being the answer is extracted. 

\section{XSense KB Construction}
\label{sec:kb}
Here we present our technique for creating a \xsensekb\ from a corpus of reviews.
Our goal is to understand what conclusions a certain expressed opinion entails.
For instance, \phrase{fresh sashimi} often implies a \phrase{good Japanese place}, but
building such knowledge bases is not trivial for several reasons. First, these
relationships are rarely mentioned explicitly in reviews. Moreover, such relationships,
while generally true, are not completely factual as there could also be a \phrase{low
quality Japanese restaurant} that serves \phrase{fresh sashimi}. Despite these
challenges, we show how the unique structure of review corpora enables us to
mine these relationships effectively.

We start by applying the opinion extractor to obtain  all (modifier, aspect) from the reviews.  We then 
create two
representations of the data:

\spara{Extraction Matrix:} We create a matrix $M$ where each row $i$ corresponds
to a product or service $i$ being reviewed and each column $j$ corresponds to a
unique (modifier, aspect) pair extracted by the opinion extractor. Each entry $M_{ij}$
denotes the number of times that the (modifier, aspect) pair $j$ has been observed in
reviews of item $i$.

\spara{Modifier-Aspect Tensor:} In a similar manner, we create a tensor $\mathcal{T}$
with three dimensions corresponding to the items, the modifiers, and the aspects
extracted from the reviews. Each entry $\mathcal{T}_{ijk}$ denotes the number of
times that modifier $j$ on aspect $k$ has been observed in reviews of item $i$.

\begin{figure}
    \centering
    \includegraphics[scale=.3]{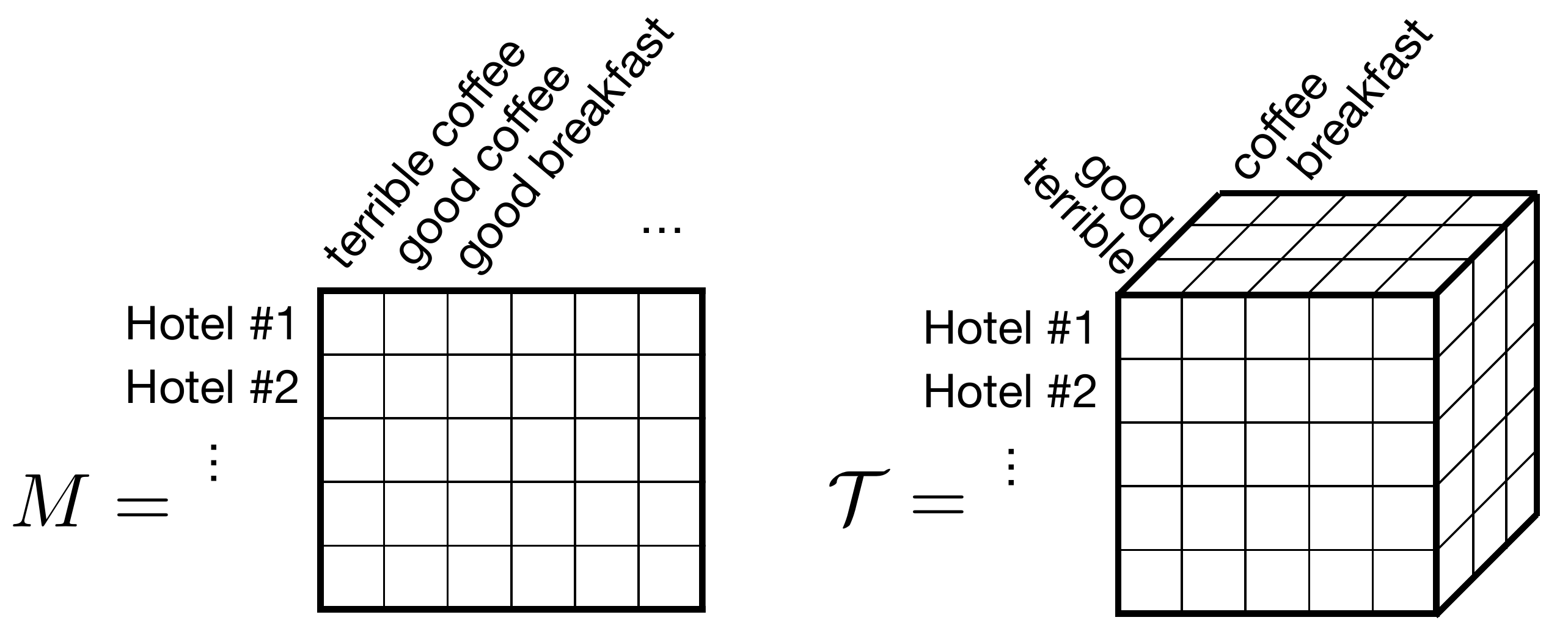}
    \caption{Representations used for KB construction}
    \label{fig:tensors}
\end{figure}

Figure~\ref{fig:tensors} illustrates how the reviews and all extracted (modifier, aspect)
pairs are organized. Using these data representations,
we compute a dense representation for each modifier-aspect pair using \emph{tensor factorization}
techniques as follows. To decompose matrix $M$, we represent each item $i$ and each
(modifier, aspect) pair $j$ with $d$-dimensional vectors $\mathbf{v}_i$ and $\mathbf{v}_j$ such that their inner product, denoted as $\widehat{M}_{ij} = \mathbf{v}_i \mathbf{v}_j$, would be a good approximation of $M_{ij}$. More specifically, we compute these vectors such that $||\widehat{M}_{ij} - M_{ij}||$ is minimized\footnote{$||.||$ denotes the Frobenius norm.}.
To decompose the modifier-aspect tensor $\mathcal{T}$ we follow a similar approach
and assign $d$-dimensional vectors $\mathbf{v}_i$, $\mathbf{v}_j$, and $\mathbf{v}_k$
to each item $i$, modifier $j$, and aspect $k$ such that the sum of their
Hadamard products, denoted as $\widehat{\mathcal{T}}_{ijk} = \mathbf{1}.(\mathbf{v}_i \circ \mathbf{v}_j \circ \mathbf{v}_k)$ would be a good approximation of $\mathcal{T}_{ijk}$. As before, we compute these representation vectors such that 
$||\widehat{\mathcal{T}}_{ij} - \mathcal{T}_{ij}||$ is minimized. These vectors are computed using a PARAFAC factorization technique~\cite{harshman1994parafac} and
we use the implementation provided by Tensorly\footnote{\url{http://tensorly.org}}.

\begin{figure}
    \centering
    \includegraphics[scale=.39]{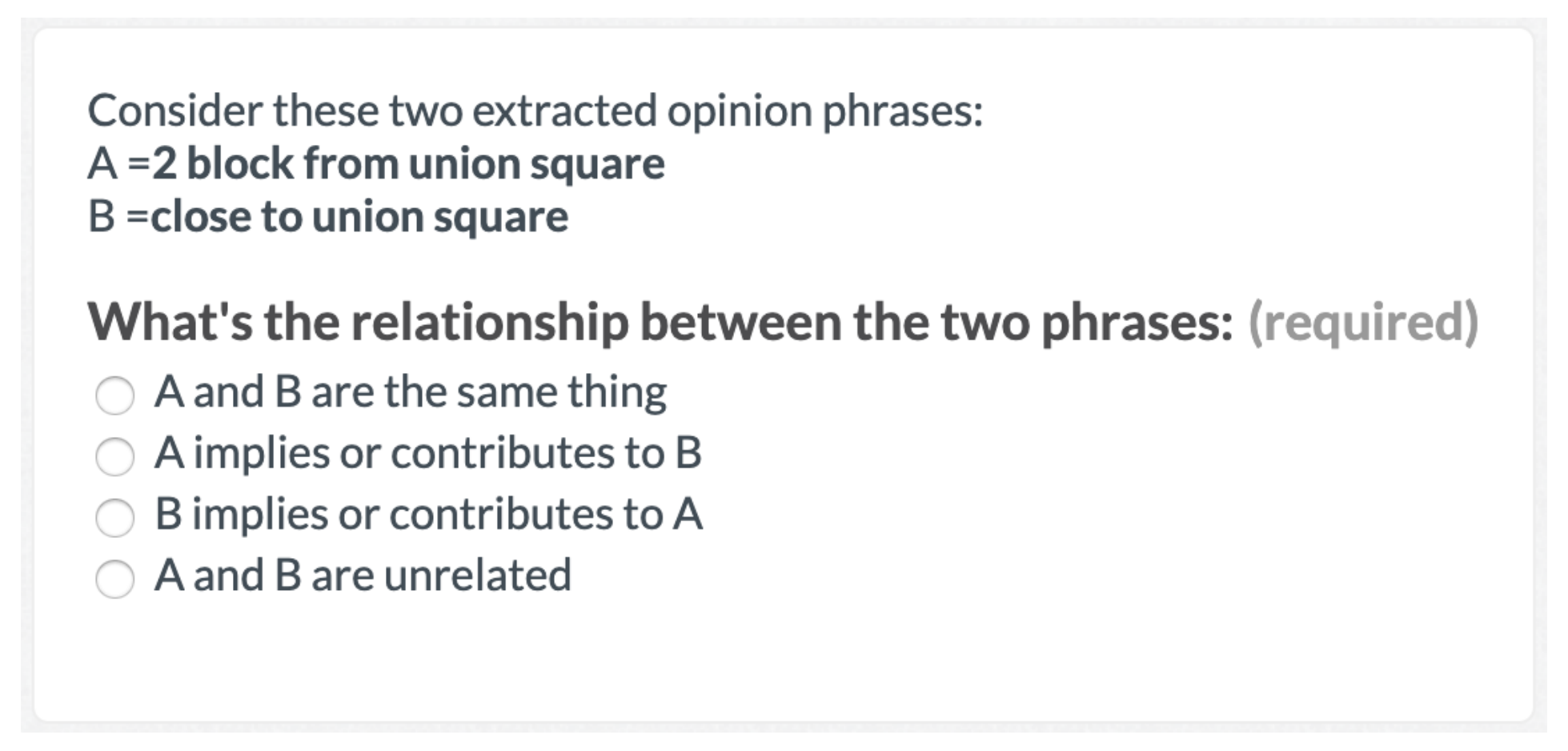}
    \caption{An example of the candidate verification task.}
    \label{fig:verify}
\end{figure}

Note that decomposing the modifier-aspect tensor produces representations for each 
modifier and aspect separately. To obtain a representation for the pair consisting of
modifier $j$ and aspect $k$, we use their Hadamard product (i.e., $\mathbf{v_j} \circ \mathbf{v_k}$). Once dense representations for all (modifier, aspect) pairs are computed,
we create a commonsense KB through the following two steps:\\
\textbf{Candidate Generation}:
In this step, we create a set of candidate premise-conclusion pairs.
More specifically, for each (modifier, aspect) pair $p$, we find $3$
other (modifier, aspect) pairs whose representations have the highest cosine similarity
with that of $p$. Also to ensure that candidate premises and conclusions are different
enough, we also find the most similar embedding with a distinct 
modifier and aspect from the pair $p$. Note that pairs with similar representations
are pairs that appear with similar distribution across all items, and thus are quite
likely to be related. The candidates mined in this step are then forwarded
to human annotators for verification.\\
\textbf{Verification}: In this step, the annotators receive a pair of extractions
and are asked to identify if the pair is \emph{unrelated}, \emph{equivalent},
or if one implies the other. Figure~\ref{fig:verify} shows an instance of our
verification task and how it was shown to human annotators.

Note that we can use either the modifier-aspect tensor or the extraction matrix for
creating the \xsensekb. However, we use both data
structures in conjunction as we observed that the extraction matrix
yields better results for frequent (modifier, aspect) pairs, and the modifier-aspect
tensor produces better results for (modifier, aspect) pairs in the long tail.
Thus by combining the results from both structures we achieve a good set of
candidates across the board.

\section{Datasets and Evaluation}
\label{sec:experiment}
In this section, we present our evaluation setting, introduce the datasets used including
our new adversarial QA dataset for the \emph{hospitality} domain, and discuss the performance of our
system \systemname compared to a number of baselines. Our experiments demonstrate two key results:
(1) {\xsensekb}s contain commonsense information 
that cannot
be derived from ConceptNet, the most popular commmonsense KB and (2) 
\xsensekb{}s improve review comprehension;
\systemname, which utilizes {\xsensekb}s, outperforms the state-of-the-art
models on multiple review comprehension tasks. To facilitate future 
research, we are making all three constructed {\xsensekb}s, and our adversarial QA
dataset publicly available online\footnote{\url{https://github.com/xsense2020/xsense}}.

\subsection{Constructed {\xsensekb}s}
We have created
three {\xsensekb}s\footnote{RestaurantSense, LaptopSense, and HospitalitySense} for improving review comprehension.
Table~\ref{table:kbs}
shows the overall statistics of the collected KBs. The first two rows denote the corpora
and the specific subset of the data that were used for creating each \xsensekb.
Once all modifier-aspect pairs were extracted from the reviews, we picked a subset
of the most reviewed entities as well as a subset of the most frequent extractions to form
the extraction matrix as well as the modifier-aspect tensor as described in Section~\ref{sec:kb}.
The number of selected entities and extractions are listed in the third row. The 
next two rows show the final number of opinions in the knowledge-base and the number
of relationships discovered between them accordingly. The last two rows in the table demonstrate
to what extent the contents of our constructed knowledge-base can be obtained from ConceptNet.
The \emph{extraction overlap} is the percentage of extracted opinions that can be directly found in
ConcpetNet. For instance, while \phrase{thin walls} appears in ConceptNet, most extracted
opinions such as \phrase{noisy room} are missing. The relation overlap denotes to what extent
the facts in our {\xsensekb}s\ can be derived \emph{indirectly} from ConceptNet. Of course, since
\phrase{noisy room} is absent from ConceptNet, we cannot derive its relationship to
\phrase{thin walls} directly. Instead, we look to see if there is a relation in
ConceptNet between the modifiers of each premise and conclusion as well as their aspects.
In this case, while there is an edge between \phrase{walls} and \phrase{rooms}, there is still no relation connecting \phrase{noisy} to \phrase{wall} in ConceptNet. 

\setlength{\tabcolsep}{2.5pt}
\begin{table}[!ht]
\small
\resizebox{\linewidth}{!}{
\begin{tabular}{cccc}\toprule
\textbf{Domain}                      & \textbf{Restaurants} & \textbf{Laptop}      & \textbf{Hospitality}   \\ \midrule
Corpus                     & Yelp        & AmazonQA    & TripAdvisor   \\
Category                      & Toronto     & Electronics & San Francisco \\
\#Ent / \#Ext                 & 2,000 / 5,000 & 2,000 / 2,000   & 860 / 2,000     \\ \midrule
\#Unique opinions          & 3,017        & 1,815        & 1,768          \\
\#Facts                    & 7,546        & 6,867        & 6,776          \\
Extraction Overlap            & 7\% & 8\% & 6\% \\
Relation Overlap            & 22\% & 24\% & 32\% \\
\bottomrule
\end{tabular}}
\caption{\small RestaurantSense, LaptopSense, HospitalitySense.}\label{table:kbs}
\end{table}


\subsection{Commonsense KB evaluation datasets}
To measure the value of using commonsense for review comprehension, we
evaluate \systemname\ on two public aspect-based sentiment analysis (ABSA)
dataset where each consists of an aspect extraction (AE) task
as well as an aspect sentiment classification (ASC) task.
Moreover, we create a QA dataset for the hospitality domain which 
is more challenging than existing QA datasets for reviews
(as shown by the low F1 scores achieved by the state-of-the-art systems).
This is because existing QA datasets for reviews are often constructed
by matching reviews with questions using IR techniques
and consequently, questions and answer spans tend to exhibit a large
similarity. 

We discuss next how our collected dataset avoids this bias, and then
describe briefly the public datasets that are used in our experiments.

\mpara{HotelQA dataset.}
We created an adversarial QA dataset (HotelQA) of 757 data entries where each question requires
commonsense reasoning in the hospitality domain to answer. Similar to \squad \cite{Rajpurkar:2016:SQuAD}, 
each data entry of HotelQA is a tuple ({\it review}, {\it question}, {\it answer})
where {\it answer} is a sentence span within {\it review}.
On average, each review has 138.6 words, each question has 5.8 words, and each answer has 19.6 words. The dataset is more challenging because we ensured questions regarding a specific topic (e.g., parking) should be paired with reviews that mention the same topic at least three times -- i.e., it is adversarial towards machine learning models -- but the concrete sub-topic (e.g., parking fee) is not mentioned explicitly in the review -- i.e., it requires commonsense reasoning to answer correctly. 
An example QA tuple is ({\it Review:} {\small ``...The best was the pre-paid parking. I booked on Expedia and included parking. A great deal! Parking was just behind the hotel and connected ...''}, {\it Question:} {\small ``Do you have parking nearby?''}, {\it Answer:} {\small ``Parking was just behind the hotel and connected.''}) 
The HotelQA dataset is separated into a training set of 681 QA pairs (90\%) and a validation set of 76 QA pairs (10\%). 


\smallskip
\noindent
\textbf{ABSA datasets. } We evaluate \systemname\ on four ABSA datasets.
The datasets cover two domains (laptops and restaurants) and consist of two tasks, AE and ASC. 
All four datasets are from SemEval competitions \cite{pontiki2014semeval,absa}.
Table \ref{tab:absa-data} summarizes the statistics of these datasets.
We split the datasets into training/validation sets following the settings
of \cite{Hu2019}, where 150 training examples are held for validation.

\setlength{\tabcolsep}{4pt}
\begin{table}[!ht]
\small
    \centering
\vspace*{-2mm}
    \resizebox{\linewidth}{!}{
    \begin{tabular}{ccc}\toprule
  Domain  & \textbf{AE} & \textbf{ASC} \\ \midrule
\textbf{Restaurant} & SemEval16 Task 5 & SemEval14 Task 4 \\ 
Train & 2000 S / 1743 A & 2164 P / 805 N / 633 Ne \\
Test & 676 S / 622 A & 728 P / 196 N / 196 Ne  \\ \midrule
\textbf{Laptop} & SemEval14 Task 4 & SemEval4 Task 4\\ 
Train & 3045 S / 2358 A & 987 P / 866 N / 460 Ne  \\
Test & 800 S / 654 A & 341 P / 128 N / 169 Ne \\ \midrule
    \end{tabular}}
    \caption{\small Statistics for the ABSA datasets. S: number of sentences; 
    A: number of aspects; P, N, and Ne: number of positive, negative and neutral polarities.} 
    \label{tab:absa-data}
\vspace*{-2mm}
\end{table}


\subsection{Experimental setup}
Next, we describe our experimental setup for each task and describe
the baseline methods. To demonstrate the importance of building
domain-specific commonsense knowledge-bases, we compare our results
with an adapted version of \systemname\ that uses embeddings from
ConceptNet, which we refer to as \systemnamecn.
We start with a description of this baseline (which we use in all
experiments), and then continue introducing our task-specific
baselines and experimental setup.

\mpara{xSense(CN). } \systemnamecn\ operates in the same manner
as \systemname. The only important change is that the
commonsense embeddings do not come from our commonsense reasoning
model. Instead, we obtain the embeddings by applying a KB embedding
technique to ConcpetNet. More specifically, we took the English
subset of ConceptNet with 873K entities and 1.5M relations, and
then embed them using the {\sc DistMult}~\cite{Yang:2015:BILINEAR}
technique for KB embedding. We use OpenKE\footnote{\url{https://github.com/thunlp/OpenKE}}
and their proposed default configuration to train the model. 

\mpara{HotelQA setup and baselines.} 
For this task, \systemname\ is implemented on top of a BERT QA model fine-tuned on \squad. 
This QA model uses a pre-trained BERT-large model~\cite{Wolf2019HuggingFacesTS} of 24
layers with 110M parameters. \systemname\ extends the original 1,024-dimensional representation
with a 768-dimensional KB vector. We compare \systemname to a baseline \hsensebase which is a
BERT model using the same configuration as \systemname. The other baseline is \systemnamecn as described above. All models are trained on the training dataset for 10 epochs with a learning rate of 3e-6 and evaluated on the evaluation dataset. We train each model 5 times and report the
average and the standard deviation of the best F1 and exact-matching scores of each run.

\mpara{ABSA setup and baselines.} 
We compare \systemname\ with BERT-PT \cite{Hu2019}, 
the SOTA method for AE and ASC. BERT-PT improves the vanilla BERT model
by concurrently fine-tuning the 12-layer BERT-based model on an in-domain
corpus and on a reading comprehension dataset. 
We reproduce the results of BERT-PT by fine-tuning the same BERT model on in-domain corpora -- 1.17 million sentences \cite{amazon} for the laptop domain and 
2 million sentences \cite{yelp} for the restaurant domain.
We use the resulting models (denoted as in-domain BERT) as a baseline
which already has similar or even better performance compared to BERT-PT.

We also use \systemname\ and \systemnamecn\ to incorporate KB embeddings.
Note that since AE and ASC are part of the opinion extraction pipeline,
we avoid the interference of having a too powerful opinion extractor 
by assuming a much weaker extractor: 
it simply takes all aspect/modifier tokens 
that appear in the \xsensekb\ as the opinion.


For all models trained based on the ABSA datsets, we fine-tune BERT for 20 epochs
with a learning rate of 5e-5. We select the model with the best performance
(F1 for AE and MF1 for ASC) on the validation set and report the performance on
the test set. We repeat each experiment 5 times and report the average.


\subsection{HotelQA and ABSA results}\label{sec:results}

\mpara{HotelQA results.} Table~\ref{exp:qa} shows the results of comparing \systemname\ with the two baselines. We report (1) the token-wise F1 scores which measure the overlap between the predictions with the golden answer and (2) the exact-matching scores -- the percentage of predictions that match exactly.
In Table \ref{exp:qa}, \systemname\ improves the base QA model by a significant 2.5\%
and by 1.3\% more compared to ConceptNet. 
We inspected the output of each model and show an example QA 
where \systemname\ outperforms the baseline models in Table~\ref{exp:qa_examples}.

\setlength{\tabcolsep}{3.5pt}
\begin{table}[t!]
\small
\centering
    \begin{tabular}{cccc} 
    \toprule
    {\bf Model} & {\bf F1 score} & {\bf Exact}\\ \midrule
    \hsensebase & 60.25 $\pm$ 0.74  & 40.26 $\pm$ 1.34 \\ 
    \systemnamecn & 61.48 $\pm$ 1.54  & 41.78 $\pm$ 1.54 \\ 
    \systemname  & 62.76 $\pm$ 1.84 & 40.79 $\pm$ 2.20\\ \bottomrule
\end{tabular}
\caption{\small Results on HotelQA with standard deviation.}
\label{exp:qa}
\end{table}

\begin{table*}[ht!]
\small
\centering
    \begin{tabular}{|p{16cm}|} 
    \hline
    {\bf Question:} Is there plenty of parking?\\
    {\bf Review:} ... The size of the bathroom is the only downside I've found. Very small! However, plenty of hot water (showerhead was not working very effectively...) and very clean bathroom and towels. The morning receptionist (I forgot to ask him his name, but I thank him again) was very nice and accepted to keep our car in the hotel parking until 3:00 p.m. at no charge. This allowed us to go shopping on St. Denis Street without being forced to find a pay parking or to run to a parking meter every 2 hours...\\
    {\bf Golden Answer:} This allowed us ... without being forced to find a pay parking or to run to a parking meter every 2 hours.\\
    \hline
    {\bf BERT+SQuAD:} However, plenty of hot water ... and very clean bathroom and towels.\\
    {\bf xSense(CN):} However, plenty of hot water ... and very clean bathroom and towels.\\
    {\bf xSense:} without being forced to find a pay parking or to run to a parking meter every 2 hours.\\
    \hline
\end{tabular}
\caption{\small An example QA pair where \systemname outperforms baselines. It is likely that the baseline models picked the ``However ...''
span because it contains ``plenty of'' which also appears in the question. xSense avoids this span perhaps because its ``bathroom'' concept
was strengthened by the commonsense vector of (``very clean'', ``bathroom'').}
\label{exp:qa_examples}
\end{table*}

\setlength{\tabcolsep}{4pt}
\begin{table*}[!ht]
    \begin{minipage}{.55\linewidth}
\small
\centering
\begin{tabular}{ccccccc} \toprule
                & \multicolumn{3}{c}{Restaurant} & \multicolumn{3}{c}{Laptop} \\
                & F1      & P  & R  & F1    & P & R \\ \midrule
BERT-PT         & 77.97   & -          & -       & 84.26 & -         & -      \\ \midrule
in-domain BERT  & 79.56	& 76.21	& 83.53	& 84.27	& 84.30	& 84.25      \\
\systemnamecn & 78.57	& 79.11	& 78.07	& 83.86	& 83.36	& 84.40      \\
\systemname     & \textbf{80.39}   & 79.66      & 81.24   & \textbf{84.44}	& 84.10	& 84.83      \\ \bottomrule
\end{tabular}
    \end{minipage}%
    \begin{minipage}{.45\linewidth}
\small
\centering
\begin{tabular}{ccccc}
\toprule
                & \multicolumn{2}{c}{Restaurant} & \multicolumn{2}{c}{Laptop} \\
                & Acc            & MF1           & Acc          & MF1         \\ \midrule
BERT-PT         & 84.95          & 76.96         & 78.07	    & 75.08          \\ \midrule
in-domain BERT  & 85.82          & 79.15         & 78.90        & 74.35      \\
\systemnamecn & 86.48	& 79.98	& 77.68	& 72.77           \\
\systemname     & \textbf{86.48}          & \textbf{80.14}         & \textbf{79.66} &	\textbf{75.76}           \\ \bottomrule
\end{tabular}
\label{exp:aspbsa}    \end{minipage} 
\caption{\small Aspect Extraction (AE, left) and Aspect Sentiment Classification (ASC, right) results. The BERT-PT numbers are taken from \cite{Hu2019}. P: precision, R: recall, MF1: Macro-F1. The standard deviation is 1.51 for AE and 1.06 for ASC.}\label{tab:absa}
\end{table*}

\mpara{ABSA results. } We summarize the results on the four ABSA datasets in Table \ref{tab:absa}.
We measure the model performance on AE tasks using F1 and the model performance
on ASC tasks using both accuracy and macro-F1.
\systemname consistently outperforms BERT-PT (SOTA) on all datasets.
The improvements range from 0.18 (F1 for Laptop AE) to 3.18 (MF1 for Restaurant ASC).
The improvement is higher for the restaurant domain.
Intuitively, this is because the restaurant KB is of better quality.
Moreover, we notice that ConceptNet hurts the BERT performance
on most cases while \systemname\ improves the baseline model
both more significantly and consistently.
These results clearly show that the domain-specific knowledge captured by the \xsensekb
is beneficial to ABSA tasks.

\section{Related Work}\label{sec:related}

\mpara{Commonsense Reasoning Tasks.} 
Several commonsense reasoning tasks have been proposed: SWAG~\cite{Zellers:2018:SWAG}, CommonsenseQA~\cite{Talmor:2019:CommonsenseQA}, and Cosmos QA~\cite{Huang:2019:CosmosQA} are multiple-choice QA tasks, and ReCoRD~\cite{Zhang:2018:ReCoRD} is a cloze-style QA task. Those datasets were carefully curated to exclude easy questions that text processing systems can answer by exploiting lexical heuristics.
To the best of our knowledge, 
HotelQA is the first span-extraction QA dataset for commonsense reasoning that has been published to date.

\mpara{External Knowledge Integration. }
A popular approach to integrating KBs into NN models is to integrate embeddings
obtained from the KB into the model. KB-LSTM~\cite{Bishan2017} incorporates
external knowledge by adding knowledge embeddings obtained from WordNet into RNN-LSTM. \citet{Yang:2019:EnhancingPretrained} applied a similar idea to BERT.
%
\citet{Mihaylov:2018:KnowledgeableReader} used an attention mechanism to integrate relevant external knowledge for cloze-style reading comprehension. \citet{Lin:2019:KagNet} developed a method that uses schema graph construction for KB embedding. 
%
All said techniques require a {\it KB-retrieval} function to find corresponding information from the KBs. 


Other approaches use an auxiliary model or data as additional evidence to the main model. \citet{Emami:2018:KnowledgeHunting} collected texts from the web using a query augmented from the input text to improve the performance on Winograd Schema Challenge (WSC). \citet{Rajani:2019:ExplainYourself} created a dataset of explanations and developed a framework that uses a language model trained to generate an {\it explanation} as an auxiliary input to the main QA model.

Our framework is different from the two approaches above. It uses a general-purpose opinion extractor and a seq2seq model that can take any input, including ones that do not explicitly appear in the KB. In contrast to the second approach, our framework directly
integrates the auxiliary information into the model, and does not provide it
as part of the input text.

\mpara{Automated KB Construction.} 
Our automatic KB construction approach is closely related to Universal Schema~\cite{yao2013universal,verga2015multilingual}, which is a matrix factorization technique for relation 
extraction. Other matrix factorization techniques for KB construction include \cite{nickel2011three} and \cite{he2015knowledge}.
A major difference, however, is that we also use tensor factorization to model aspects and modifiers separately.

\noindent
{\bf Review Comprehension}
\citet{Hu2019} introduced the Review Reading Comprehension task, and created a new reading comprehension dataset based on crowd-sourced questions on reviews in the ABSA datasets~\cite{pontiki2015semeval}. 
We demonstrate that \systemname{} outperforms their approach and achieves the SOTA results (see Table~\ref{tab:absa}).


\section{Conclusion}
\label{sec:conclusion}

We establish that domain-specific commonsense 
can noticeably improve multiple review comprehension tasks that conventional commonsense knowledge bases cannot. 
We develop \systemname, a system that can exploit relatively small domain-specific knowledge bases on top of transformer-based language models for review comprehension and establish its effectiveness through an extensive set of experiments. To facilitate further research, we also
publicly release three domain-specific knowledge bases, in the domains of hospitality, restaurant, and laptops, and release a question-answering benchmark for the hospitality domain.

\bibliography{refs}
\bibliographystyle{acl_natbib}

\end{document}